\documentclass[10pt,twocolumn,letterpaper]{article}

\usepackage{wacv}              

\usepackage{times}
\usepackage{epsfig}
\usepackage{graphicx}
\usepackage{amsmath}
\usepackage{amssymb}

\usepackage{booktabs}
\usepackage{kotex}
\usepackage{multirow}
\usepackage{makecell}
\usepackage{color, colortbl}
\usepackage{caption}
\usepackage[accsupp]{axessibility}  
\usepackage[pagebackref,breaklinks,colorlinks]{hyperref}

\usepackage[capitalize]{cleveref}
\crefname{section}{Sec.}{Secs.}
\Crefname{section}{Section}{Sections}
\Crefname{table}{Table}{Tables}
\crefname{table}{Tab.}{Tabs.}

\begin{document}
\def\MethodName{UGPNet}
\newcommand{\XX}{{\textcolor{red}{XX}}}

\newcommand{\Eq}[1]   {Eq.\ (#1)}
\newcommand{\Eqs}[1]  {Eqs.\ (#1)}
\newcommand{\Fig}[1]  {Fig.\ #1}
\newcommand{\Figs}[1] {Figs.\ #1}
\newcommand{\Tbl}[1]  {Tab.\ #1}
\newcommand{\Tbls}[1] {Tabs.\ #1}
\newcommand{\Sec}[1]  {Sec.\ #1}
\newcommand{\SSec}[1] {Sec.\ #1}
\newcommand{\Secs}[1] {Secs.\ #1}
\newcommand{\Alg}[1]  {Alg.\ #1}
\renewcommand{\etal}    {\textit{et al.}}

\title{\MethodName: Universal Generative Prior for Image Restoration}

\author{Hwayoon Lee$^{1, \ast}$ \quad Kyoungkook Kang$^{2}$ \quad Hyeongmin Lee$^{2}$ \quad Seung-Hwan Baek$^{2}$ \quad Sunghyun Cho$^{2}$\\
\and
$^1$GENGENAI\\
{\tt\small hwayoon.lee@gengen.ai}
\and
$^{2}$POSTECH\\
{\tt\small \{kkang831, hmin970922, shwbaek, s.cho\}@postech.ac.kr}
}

\twocolumn[{
\renewcommand\twocolumn[1][]{#1}
\maketitle
\begin{center}
    \centering
    \captionsetup{type=figure}
    \includegraphics[width=0.98\textwidth]
    {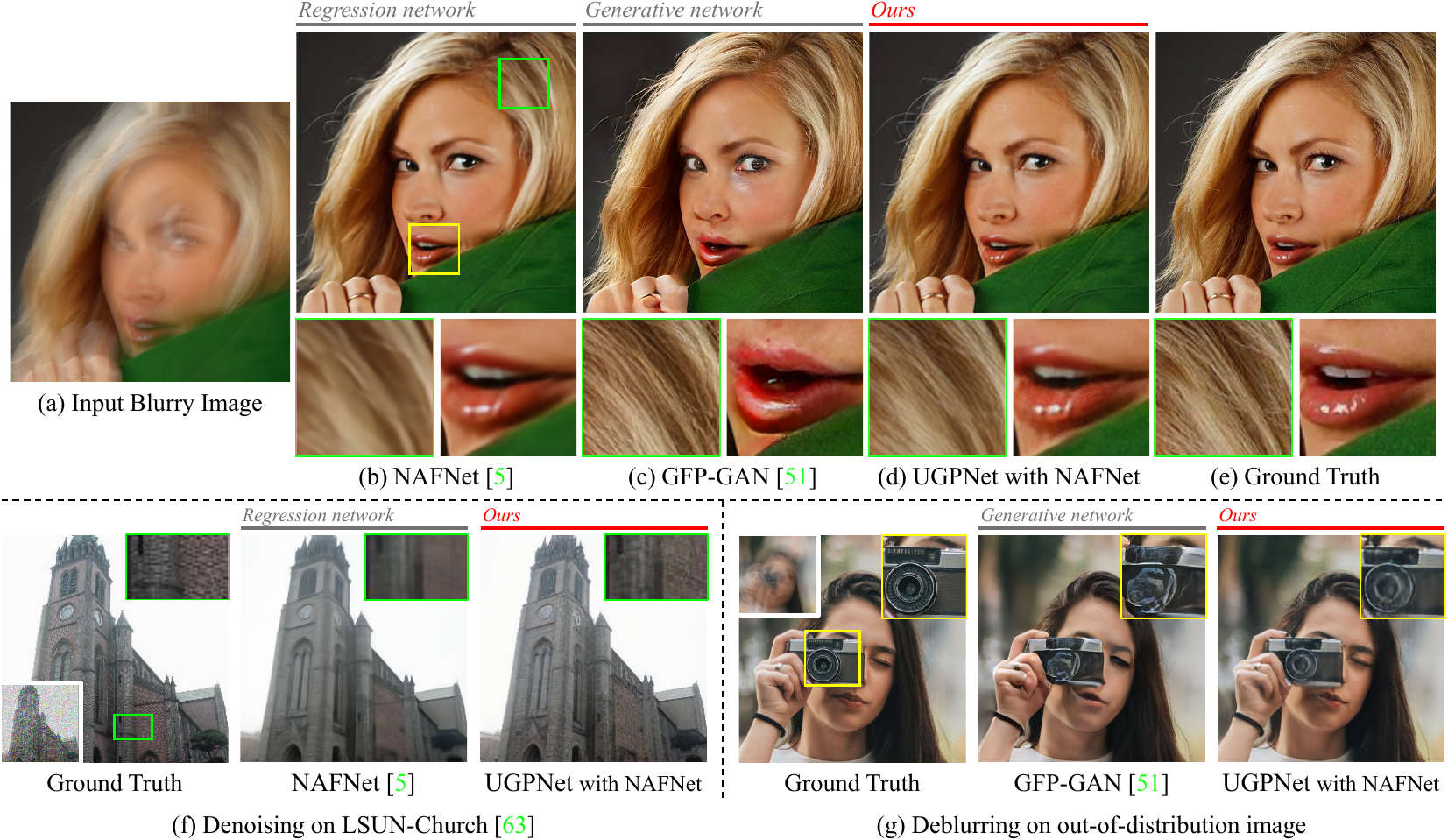}
    \vspace{-0.2cm}
    \captionof{figure}{
    We present \MethodName{}, a universal image restoration framework that combines the benefits of an existing regression-based restoration network and a generative prior-based network.
    (a) Given degraded images, e.g. a blurry one, (b) a regression network~\cite{NAFNet} fails to recover perceptually-realistic details while it recovers the coarse structure of the original image.
    (c) In contrast, a generative network~\cite{gfpgan} synthesizes perceptually-realistic high-frequency details while sacrificing structural consistency with the input image. (d) \MethodName{} allows us to maintain the original structure of the input image and synthesize perceptually-realistic high-frequency details.
    As a universal framework, (f) \MethodName{} is applicable to natural images~\cite{lsun}. In addition, (g) it is robust against catastrophic failures that generative prior-based methods encounter when restoring images outside the training distributions.
    }
    \label{fig:teaser}
\end{center}
}]
{\let\thefootnote\relax\footnotetext{\noindent ${}^{\ast}$This work was done at POSTECH.}}

\begin{abstract}
\vspace{-0.3cm}
Recent image restoration methods can be broadly categorized into two classes: (1) regression methods that recover the rough structure of the original image without synthesizing high-frequency details and (2) generative methods that synthesize perceptually-realistic high-frequency details even though the resulting image deviates from the original structure of the input.
While both directions have been extensively studied in isolation, merging their benefits with a single framework has been rarely studied.
In this paper, we propose \MethodName, a universal image restoration framework that can effectively achieve the benefits of both approaches by simply adopting a pair of an existing regression model and a generative model.
\MethodName{} first restores the image structure of a degraded input using a regression model and synthesizes a perceptually-realistic image with a generative model on top of the regressed output.
\MethodName{} then combines the regressed output and the synthesized output, resulting in a final result that faithfully reconstructs the structure of the original image in addition to perceptually-realistic textures.
Our extensive experiments on deblurring, denoising, and super-resolution demonstrate that \MethodName{} can successfully exploit both regression and generative methods for high-fidelity image restoration.
\vspace{-0.4cm}
\end{abstract}

\section{Introduction}
\vspace{-0.1cm}
Image restoration~\cite{fastmotiondeblurring,non-localdenoise,collaborativefiltering,removingshake} has been studied for decades in computer vision and graphics.
To recover natural structure from unwanted image degradation such as blur, noise, and low resolution, various non-learning and learning-based approaches have been proposed. 

Since the advent of deep neural networks, \emph{regression methods}~\cite{hinet,mprnet_multiple,mimounet,NAFNet,restormer_multiple,uformer_multiple,maxim_multiple} have emerged as an effective approach for image restoration.
Regression methods commonly use convolution neural networks and directly map an input degraded image to a clean image via regression losses such as mean-squared error (MSE). 
While faithful reconstruction of the original image structure can be achieved, regression methods tend to enforce the resulting image to follow the average image of all potential natural images corresponding to the input. This results in a blurry image without high-frequency details, which harms the perceptual realism of restored images. See \Fig{\ref{fig:teaser}} (b) and (f).

\emph{Generative approach} is another major principle for image restoration that tackles the high-frequency detail problem.
Learning natural image statistics using generative models~\cite{stylegan2,stylegan1,vqgan} allows synthesizing perceptually-realistic high-frequency image details from degraded input images~\cite{gfpgan,gpen,gcfsr,glean,VQFR}. 
Exploiting the latent space of generative models via inversion~\cite{image2stylegan,pSp,e4e} further improves the naturalness of synthesized  details. Generative methods, however, heavily depend on the synthesis ability of their generators, which is often limited in recovering the exact structure of the input images. As a result, they suffer from the deviation of synthesized image structures from the original image, resulting in the loss of the identity of the portrait image (\cref{fig:teaser} (c)) or severe artifacts (\cref{fig:teaser} (g)).

Another important limitation of the existing generative approaches is their limited extendability,
which prevents them from enjoying the rapid advance of regression-based methods.
The existing generative approaches rely on their own fixed network architectures tightly coupled with generative priors.
Thus, it is hard to extend them for other restoration tasks such as deblurring as will be shown in \cref{sec:experiment} or to combine with other restoration networks.

In this paper, we present \MethodName, a universal generative prior framework for image restoration that can enjoy the restoration power of state-of-the-art regression-based methods and perceptually-realistic high-frequency details from generative priors.
\MethodName{} is designed as a flexible framework that can plug-and-play an arbitrary regression-based image restoration module.
Thanks to its flexibility, we can replace the restoration module with a more suitable one for different tasks or a more effective architecture in the future.

\MethodName{} is designed with simplicity and effectiveness in mind to make the framework easily adaptable to various restoration tasks and produce high-quality results.
Specifically, \MethodName{} consists of three modules: restoration, synthesis, and fusion.
The restoration module is a neural network for regression-based image restoration, whose architecture can be flexibly chosen by users.
The synthesis module plays the role of a generative prior and synthesizes high-frequency details suitable for the output of the restoration module.
Finally, the fusion module takes the outputs of both restoration and synthesis modules and produces a final result of high fidelity and high perceptual quality (\cref{fig:teaser}(d)).

We evaluate the effectiveness of \MethodName{} on multiple restoration tasks, including deblurring, super-resolution, and denoising.
We demonstrate that \MethodName{} successfully brings the generative power of a generative prior to state-of-the-art regression approaches, enabling faithful restoration of image structures as well as the synthesis of high-frequency details of high perceptual quality.
\section{Related Work}
\label{sec:related work}

\vspace{-0.1cm}
\paragraph{Regression-based Restoration Networks}
Deep neural networks have found their applications in image restoration.
Various network architectures have been proposed for each task, such as denoising~\cite{DnCNN,FFDNet,VDN,nbnet_denoise,practical_denoise,transfer_denoise,invertible_denoise,Blindspot_denoise}, deblurring~\cite{Gopro,SRN,mimounet,mssnet_deblur,Fourier_deblur,motionaware_deblur,attentive_deblur,pyramid_deblur}, and super-resolution~\cite{SRCNN,RCAN,activating_super,holistic_super,mining_super,progressive_super,context_super}, and also for multiple tasks~\cite{hinet,NAFNet,restormer_multiple,uformer_multiple,maxim_multiple,mprnet_multiple,MIRNet_multiple}.
This flood of research is still ongoing, rapidly breaking performance records every year.
On the other hand, regression-based methods commonly suffer from blurry textures.
Their regression losses that minimize the distortion between the restored and ground-truth images via a distance metric such as mean-absolute error (MAE) and mean-squared error (MSE) lead to restored results close to an average of all possible realistic images.
Unfortunately, such an average image inherently has blurry textures~\cite{srgan}.

Our proposed framework, \MethodName{}, is particularly designed to resolve the blurry texture problem of the regression-based methods by synthesizing realistic textures on top of their results using a generative prior.
At the same time, \MethodName{} is designed to be flexible to allow the plug-and-play of regression-based methods of different tasks.
Thanks to the flexibility of \MethodName{}, we can enjoy the state-of-the-art restoration quality of recent and even future regression-based methods and realistic textures.

\vspace{-0.3cm}
\paragraph{Regression Networks with Adversarial Losses}
To achieve perceptually pleasing restoration results with realistic high-frequency textures, recent methods~\cite{esrgan,srgan,realesrgan,deblurgan2,deblurgan,wavelet1,wavelet2} adopt adversarial losses with discriminators.
However, adopting only an adversarial loss without exploiting pretrained generator networks of GANs tends to produce unrealistic textures that do not fit the context compared to the synthesis approaches that leverage the network architectures and pretrained prior knowledge of existing generative models, which will be discussed in the following.

\vspace{-0.4cm}
\paragraph{Synthesis Networks with Generative Prior}
To benefit from the remarkable synthesis capability of existing generative models~\cite{stylegan1,stylegan2,biggan,vqgan}, the generative priors learned in generative models have recently been exploited, and enabled high-quality image restoration including blind face restoration~\cite{VQFR,gfpgan,gpen,sgpn}, super-resolution~\cite{gcfsr,glean}, and colorization~\cite{towardvivid,bigcolor}.
To exploit the generative prior learned in a pretrained GAN model, early works adopt a GAN inversion approach that optimizes the latent code by iteratively minimizing the discrepancy between the input and generated images in consideration of image degradation~\cite{pulse,mganprior}.
Instead of directly optimizing the latent code, recent works utilize encoder networks that estimate the latent code, which is subsequently fed to a GAN generator for synthesizing a clean image~\cite{gpen,gfpgan,gcfsr,glean,VQFR}.
These methods embed a generator into their networks and inherit their ability to synthesize realistic details.

Although these generative-prior-based methods have shown to be able to produce perceptually-realistic textures for image restoration, they suffer from limited representation power.
We propose an effective solution that introduces a generative prior on top of high-fidelity regression methods for faithful restoration.

More recently, a few works have explored the use of diffusion models~\cite{DDPM} for image restoration. 
However, these models cannot benefit from regression-based methods like other generative-prior-based methods. 
Furthermore, the slow inference speed~\cite{diffusion_deblur,diffusion_sr,DDNM,DDRM} and
the strict assumption on the degradation model~\cite{DDNM,DDRM} pose significant challenges to their practical use.
\begin{figure*}[t!]
    \centerline{\includegraphics[width=0.85\textwidth]{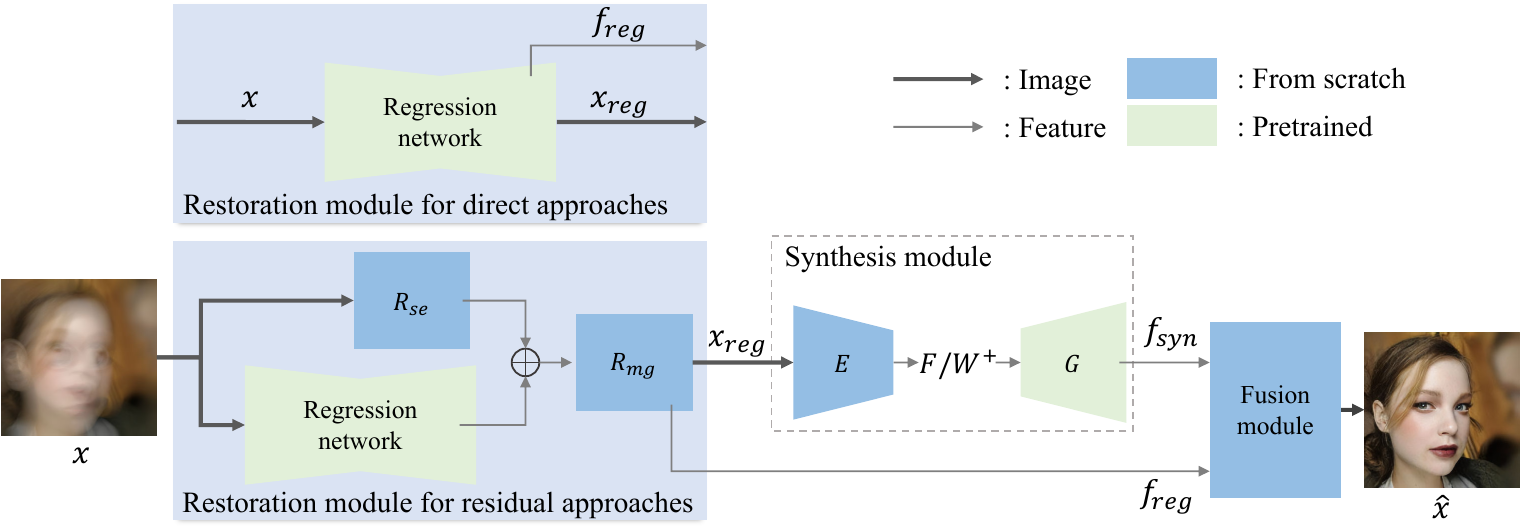}}
    \vspace{-0.1cm}
    \caption{
    \MethodName{} consists of three sub-modules: restoration, synthesis, and fusion modules.
    Given a degraded input image $x$, the restoration module first recovers the original image structure exploiting a regression network. On top of the regressed output, the synthesis module synthesizes high-frequency details exploiting a generative network. Lastly, the fusion module combines the latent features from both modules to generate a final restored image $\hat{x}$. 
    \vspace{-0.5cm}
    }
    \label{fig:framework}
\end{figure*}

\begin{figure}[t!]
\centerline{\includegraphics[width=0.8\columnwidth]{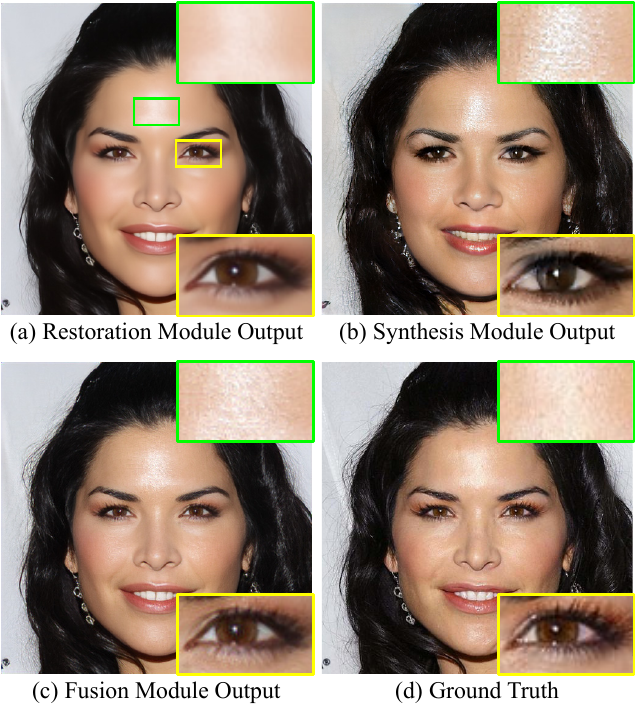}}
    \vspace{-0.2cm}
    \caption{
    An example of image outputs of \MethodName{}'s modules and corresponding ground-truth image.
    On top of the image structure of restoration module output (a) $x_{reg}$, the fusion module brings high-frequency details of synthesis module output (b) $x_{syn}$ to generate the final output (c) $\hat{x}$.}
    \label{fig:Results of each module}
    \vspace{-0.7cm}
\end{figure}

\section{\MethodName}
\label{sec:method}
\vspace{-0.1cm}

\cref{fig:framework} illustrates an overview of \MethodName{}, which consists of restoration, synthesis, and fusion modules.
\MethodName{} takes a degraded image $x$ as input, and feeds it to the restoration module.
The restoration module first recovers the original image structure exploiting an existing regression-based restoration network.
The synthesis module then synthesizes high-frequency details by inverting and regenerating the restored image through a generative network. 
The fusion module combines the latent features from the restoration and synthesis modules to generate a final restored image $\hat{x}$ that maintains both high-frequency details as well as the structure of the original input. 
In the following, we describe each module in detail.

\vspace{-0.1cm}
\subsection{Restoration Module}
\vspace{-0.1cm}
\label{subsec:restoration_module}

Given a degraded input image $x$, the restoration module aims to recover authentic image structures via an off-the-shelf pretrained regression-based restoration network.
In order to enable flexible selection among diverse regression networks, we propose simple strategies for handling two types of CNN-based regression approaches: direct and residual approaches.
Direct approaches such as RRDBNet~\cite{esrgan} directly regress the pixel values of a clean image from an input degraded image. 
In contrast, residual approaches~\cite{vdsr_residual,NAFNet,hinet}, which are gaining popularity thanks to their effectiveness in handling various restoration tasks, learn to predict a residual image that is later added to the input image to restore a clean image.

In the case of direct approaches, we adopt a direct regression network into our restoration module without any modification, and connect its output $x_{reg}$ to the synthesis module, and its last-layer feature map $f_{reg}$ to the fusion module (\cref{fig:framework}).

In the case of residual approaches, we introduce a slight modification for additional quality improvement and for the connection with the fusion module that takes features.
Specifically, a residual regression network predicts residual information and adds them to the input image in the image domain at its last stage, i.e.,
$x_{reg} = R(x) + x$
where $R(x)$ is a residual image predicted by the regression network $R$.
Instead, we modify this stage to perform in the feature domain to extract a feature map on the final output that will be fed to the fusion module.
Specifically, our restoration module obtains a restored output $x_{reg}$ as
$x_{reg} = R_{mg}( R'(x) + R_{se}(x) )$
where $R_{mg}$ is a merging network, and $R_{se}$ is a structure encoder network that embeds an input image $x$ into the feature domain.
$R'(x)$ is the last-layer feature map of $x$ estimated by $R$.

This modification enables us to use the last-layer feature map of $R_{mg}$ as the feature map $f_{reg}$ for the fusion module.
Moreover, the residual information is estimated and added back to the input image in a higher-dimensional feature space where different image characteristics can be better represented.
As a result, the restoration quality can be further improved, as will be demonstrated in \cref{sec:experiment}.

\vspace{-0.1cm}
\subsection{Synthesis Module}
\vspace{-0.1cm}
The synthesis module takes the regression result $x_{reg}$ from the restoration module and synthesizes a clean image that has similar image structures to $x_{reg}$ and realistic textures.
To this end, our synthesis module adopts the GAN inversion approach that embeds an image into the latent space of a pretrained GAN model.
Specifically, the synthesis module is composed of an encoder $E$ and a generator $G$.
The encoder estimates the latent code of $x_{reg}$ in the GAN latent space.
Then, the generator synthesizes an image with realistic textures from the latent code.
Finally, the last-layer feature of the generator $f_{syn}$ is extracted and fed to the fusion module.

For high-quality synthesis, we adopt a pretrained generator of StyleGAN2~\cite{stylegan2} for our generator $G$.
Also, to achieve a high-fidelity result that faithfully reconstructs the input restored image $x_{reg}$, our encoder $E$ embeds $x_{reg}$ into the latent space $\mathcal{F}/\mathcal{W}^+$ proposed by BDInvert~\cite{BDInvert}, which supports GAN inversion of a wide range of out-of-distribution images.
While BDInvert directly estimates the latent code in the $\mathcal{F}$ space in a feed-forward manner, it uses an iterative optimization to estimate the latent code in the $\mathcal{W}^+$ space, resulting in significant computational overhead.
To resolve this, our encoder employs an additional CNN inspired by the map2style network of pSp~\cite{pSp} to directly estimate the latent code in the $\mathcal{W}^+$ space.
The latent code in the $\mathcal{F}/\mathcal{W}^+$ space is then fed to the StyleGAN2 generator $G$ to produce a final synthesis result $x_{syn}$ and its feature map $f_{syn}$. For $f_{syn}$, we use the feature map before the last toRGB layer of the StyleGAN2 generator.

The original pSp method uses multiple map2style networks for high-fidelity reconstruction, which incurs significant overhead in terms of model size~\cite{pSp}.
On the other hand, as we employ the $\mathcal{F}/\mathcal{W}^+$ space guaranteeing coarse-level alignment and adjust the remaining spatial misalignment in the subsequent fusion module, we could share a single map2style network, which only requires approximately 10\% of the parameters without quality degradation.
Refer to the supplementary material for the detailed architecture of the encoder.

\vspace{-0.1cm}
\subsection{Fusion Module}
\vspace{-0.1cm}

{The fusion module combines the authentic image structure of $x_{reg}$ and realistic texture of $x_{syn}$ to generate the final output $\hat{x}$, as shown in \Fig{\ref{fig:Results of each module}}.
To this end, our fusion module combines the feature outputs $f_{reg}$ and $f_{syn}$ instead of image outputs $x_{reg}$ and $x_{syn}$.
This feature domain fusion helps circumvent potential misalignment between the restoration and synthesis outputs.}
We design the fusion module with residual blocks and convolution layers as:
\begin{equation}
	\hat{x} = Conv(R_{fusion}\left(Conv(f_{syn}) + f_{reg}\right)),
\end{equation}
where $R_{fusion}$ is a CNN consisting of eight residual blocks and $Conv$ denotes a $3\times3$ convolution layer.

\subsection{Training}
\vspace{-0.1cm}
We train each module of \MethodName{}  separately;
we first train the restoration module, then the synthesis module, and finally the fusion module.
In this section, we describe the training strategy for each module.
\vspace{-0.4cm}
\paragraph{Restoration Module}
In the case of direct regression methods, we use them without any modification as discussed in \cref{subsec:restoration_module}.
Thus, we can assume that they are already carefully trained to restore high-fidelity results, which leaves us nothing to train further.
On the other hand, in the case of residual regression methods, we adopt a slight modification including an additional structure encoder and a merging network.
Thus, we train the restoration module only in this case.
We train the restoration module using a loss $\mathcal{L}_{res}$ to fuse the structural information and residual information in the feature domain.
Specifically, we use the same loss function originally used to train the regression network to encourage the output of the merging network $x_{reg}$ to be close to its corresponding ground-truth image.
As \MethodName{} adopts a pretrained regression network, its weights are initialized with its own pretrained weights and further finetuned during the training. The structure encoder and the merging network are trained from scratch.
\vspace{-0.4cm}
\paragraph{Synthesis Module}
To train the synthesis module, we employ a discriminator $D$ and jointly train the encoder $E$, generator $G$, and discriminator $D$.
We initialize $G$ and $D$ using the weights of a pretrained StyleGAN2 model, while training $E$ from scratch.
The encoder and generator are trained using a loss $\mathcal{L}_{syn}$ defined as:
\begin{equation}
	\mathcal{L}_{syn} = \mathcal{L}_{1} + \lambda_{per}\mathcal{L}_{per} + \lambda_{adv}\mathcal{L}_{adv}, 
\end{equation}
where $\mathcal{L}_1$ and $\mathcal{L}_{per}$ are an $L^1$ loss and an LPIPS loss~\cite{lpips} between $x_{syn}$ and its corresponding ground-truth image, respectively.
$\mathcal{L}_{adv}$ is an adversarial loss.
$\lambda_{per}$ and $\lambda_{adv}$ are weights for $\mathcal{L}_{per}$ and $\mathcal{L}_{adv}$, respectively.
We adopt the non-saturating loss for $\mathcal{L}_{adv}$ \cite{stylegan2}.
To train the discriminator, we use a logistic loss following StyleGAN2~\cite{stylegan2}.
\vspace{-0.4cm}
\paragraph{Fusion Module}
The goals of the fusion module are twofold.
The first goal is to produce restored images that faithfully reconstruct the ground-truth clean images.
The second goal is to produce restored images with realistic textures by using the output of the synthesis module.
Based on these goals, the fusion module is trained using a loss $\mathcal{L}_{fusion}$ defined as:
\begin{equation}
	\mathcal{L}_{fusion} = \mathcal{L}_{1} + \lambda_{per}\mathcal{L}_{per} +  \lambda_{cf}\mathcal{L}_{cf},
\end{equation}
where $\mathcal{L}_1$ and $\mathcal{L}_{per}$ are an $L^1$ loss and an LPIPS loss~\cite{lpips} between the output of the fusion module $\hat{x}$ and its corresponding ground-truth image, respectively. The two loss terms are used for the faithful reconstruction of the ground-truth clean image.
$\mathcal{L}_{cf}$ is a patch-wise contextual loss~\cite{contextual} that measures the average distance between the closest feature of $x_{syn}$ for each feature of $\hat{x}$ in a patch-wise manner. $\mathcal{L}_{cf}$ transfers the textures in $x_{syn}$ to the fusion result $\hat{x}$.
As $x_{syn}$ is a synthesized image, it may have structures and details unaligned with those of $\hat{x}$. The patch-wise contextual loss handles such an alignment issue by searching the closest feature in a local patch.
The mathematical definitions of the losses are provided in the supplementary material.
\definecolor{color_mse}{rgb}{1,1,0.57}
\definecolor{color_gan}{rgb}{1,0.8549,0.5569}
\definecolor{color_ours}{rgb}{0.88,1,1} 
\section{Experiments}
\label{sec:experiment}
\vspace{-0.1cm}
\paragraph{Implementation and Evaluation Details}
{
For the evaluation of \MethodName{}, we use NAFNet~\cite{NAFNet} for denoising and deblurring, and RRDBNet~\cite{esrgan} for super-resolution as the regression network unless otherwise noted.
We train \MethodName{} on 70,000 face images of size $512\times512$ in the FFHQ dataset~\cite{stylegan1} and test on 3,000 images of the CelebA-HQ dataset~\cite{pggan}.
We synthesize a degraded version of a clean image as follows.
For denoising, we synthesize noisy images by adding Gaussian ($\mu=0$, $\sigma=0.3$) and Poisson noise ($k=30$). 
For deblurring, we apply random motion blur sampled from 1,000 motion blur kernels of size $71\times71$ synthesized following Rim~\etal~\cite{RealBlur}.  
For super-resolution, we apply $8\times$ downsampling with bicubic interpolation.
For evaluation, we use PSNR, SSIM~\cite{ssim}, LPIPS~\cite{lpips}, and FID~\cite{fid}.
PSNR, SSIM, and LPIPS measure how similar the restored image and the ground-truth image are in terms of pixel values, structural similarity, and perceptual similarity, respectively,
while FID evaluates the perceptual quality of the restored image.
More details can be found in the supplementary material.

\vspace{-0.1cm}
\begin{figure*}[ht!]
    \centerline{\includegraphics[width=2\columnwidth]{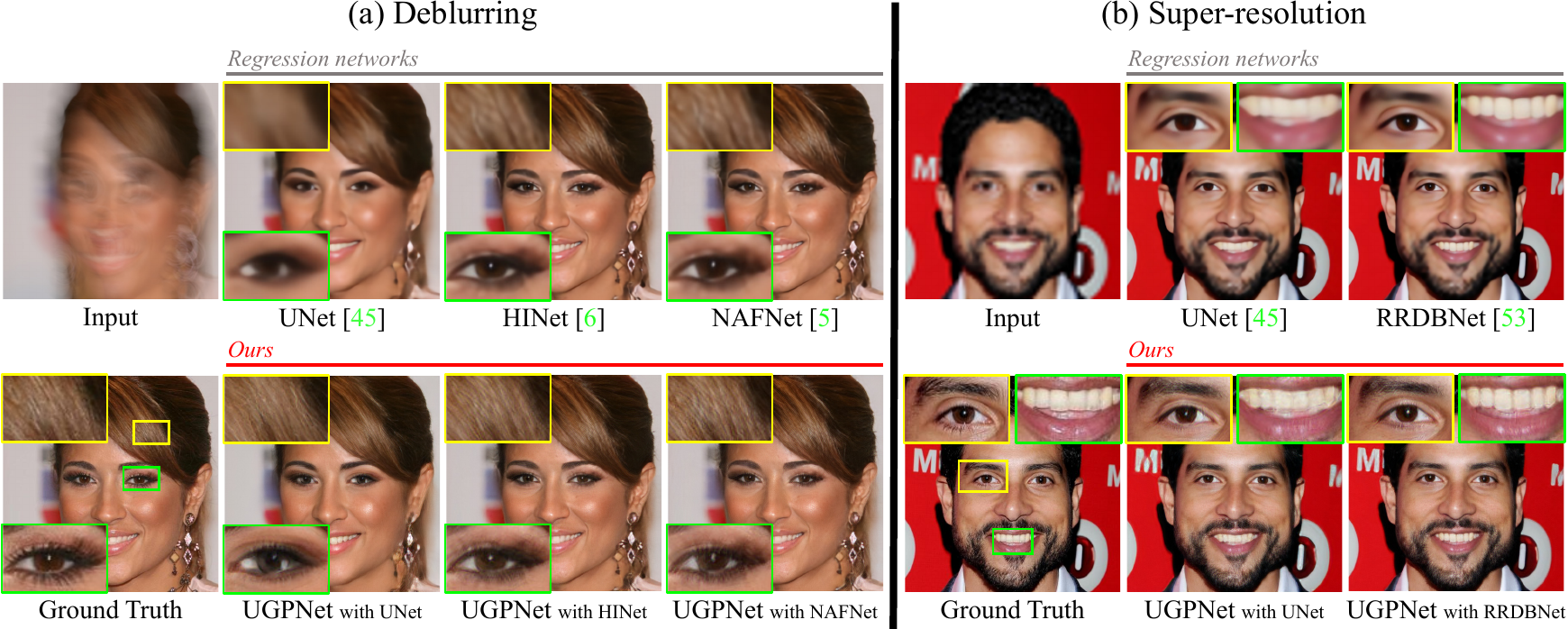}}
    \vspace{-0.2cm}
    \caption{
    UGPNet allows flexible selection of diverse regression networks in the restoration module.
    We show restoration results using regression models (UNet~\cite{UNet}, HINet~\cite{hinet}, NAFNet~\cite{NAFNet} and RRDBNet~\cite{esrgan}) on the top row for (a) deblurring and (b) super-resolution. We can equip any regression models into UGPNet that synthesizes perceptually-realistic high-frequency details, as shown in the bottom row.
    }
    \vspace{-0.2cm}
\label{fig:flexible}
\end{figure*}

\begin{table*}[t!]
        \centering
        \setlength{\tabcolsep}{0.3em}
        \scalebox{0.83}{
        \begin{tabular}{c|cccc|cccc|c|cccc}
        \Xhline{4\arrayrulewidth}
                & \multicolumn{4}{c|}{\textit{(a) Denoising}}                                                             & \multicolumn{4}{c|}{\textit{(b) Deblurring}}                   & \multicolumn{1}{c|}{}                                          & \multicolumn{4}{c}{\textit{(c) Super-resolution}}  \\ \hline
        Method  & \multicolumn{1}{c|}{\bf{PSNR~$\uparrow$}} & \multicolumn{1}{c|}{\bf{SSIM~$\uparrow$}} & \multicolumn{1}{c|}{\bf{LPIPS~$\downarrow$}} & \bf{FID~$\downarrow$} & \multicolumn{1}{c|}{\bf{PSNR~$\uparrow$}} & \multicolumn{1}{c|}{\bf{SSIM~$\uparrow$}} & \multicolumn{1}{c|}{\bf{LPIPS~$\downarrow$}} & \bf{FID~$\downarrow$} & Method  & \multicolumn{1}{c|}{\bf{PSNR~$\uparrow$}} & \multicolumn{1}{c|}{\bf{SSIM~$\uparrow$}} & \multicolumn{1}{c|}{\bf{LPIPS~$\downarrow$}} & \bf{FID~$\downarrow$} \\ 
        \Xhline{2\arrayrulewidth}
                \rowcolor{color_mse}
        NAFNet~\cite{NAFNet}  & {30.23}     & {\bf{0.82}}     & {\bf{0.30}}      &  {\bf{39.32}}   & {29.53}     & {0.81}    & {0.31}      &  42.18                                  & RRDBNet~\cite{esrgan}  & {\bf{30.03}}     & {\bf{0.80}}     & {0.32}      &  52.42   \\ \hline
                \rowcolor{color_mse}
        Uformer~\cite{uformer_multiple}  & {\bf{30.24}}     & {\bf{0.82}}     & {0.31}      &  {40.50}   & {\bf{30.38}}     & {\bf{0.83}}    & {\bf{0.29}}      &  \bf{37.02}                                   & ESRGAN~\cite{esrgan}   & {26.88}     & {0.69}     & {\bf{0.31}}      &   \bf{8.92}  \\ \hline
        \rowcolor{color_gan}
        GFP-GAN~\cite{gfpgan} & {\bf{27.96}}     & {\bf{0.76}}     & {\bf{0.31}}      &  11.66   & {\bf{23.09}}     & {\bf{0.63}}     & {\bf{0.34}}      &  \bf{10.73}                        & GFP-GAN~\cite{gfpgan}  & {27.38}     & {0.72}     & {0.28}      &   6.88  \\ \hline
        \rowcolor{color_gan}
        GPEN~\cite{gpen}    & {27.50}     & {0.73}     & {0.35}      &  11.84   & {20.60}     & {0.56}     & {0.43}      &  15.48                                                             & GPEN~\cite{gpen}     & {26.35}     & {0.69}     & {0.32}      &   9.89  \\ \hline
        \rowcolor{color_gan}
        VQFR~\cite{VQFR}    & {27.45}     & {0.74}     & {\bf{0.31}}      & \bf{11.64}    & {21.80}     & {0.59}     & {0.35}      &  10.81                                                   & VQFR~\cite{VQFR}     & {27.16}     & {0.71}     & {0.27}      &   6.41  \\ \hline
        \cellcolor{color_ours}\MethodName{}    & \cellcolor{color_ours}{\bf{29.20}}     & \cellcolor{color_ours}{\bf{0.78}}     & \cellcolor{color_ours}{\bf{0.31}}      &  \cellcolor{color_ours}\bf{10.24}   & \cellcolor{color_ours}{\bf{28.64}}     & \cellcolor{color_ours}{\bf{0.76}}     & \cellcolor{color_ours}\bf{0.31}      &  \cellcolor{color_ours}\bf{8.02}
        & \cellcolor{color_gan}GLEAN~\cite{glean}    & \cellcolor{color_gan}{27.74}     & \cellcolor{color_gan}0.72     & \cellcolor{color_gan}0.28      &   \cellcolor{color_gan}6.47  \\ \hline
           &      &    &     &    &     &     &      &                                                                                                                                                                                                                                                    
        &  \cellcolor{color_gan}GCFSR~\cite{gcfsr}    & \cellcolor{color_gan}\bf{28.16}     & \cellcolor{color_gan}\bf{0.74}    & \cellcolor{color_gan}\bf{0.26}      &   \cellcolor{color_gan}\bf{5.72} \\ \hline          
           &      &    &     &    &     &     &      &                                                                                                                                   
        &  \cellcolor{color_ours}\MethodName{}     & \cellcolor{color_ours}\bf{28.70}    & \cellcolor{color_ours}\bf{0.74}     & \cellcolor{color_ours}\bf{0.30}      &   \cellcolor{color_ours}\bf{6.70}  \\ 
        \Xhline{4\arrayrulewidth}
\end{tabular}}
\vspace{-0.3cm}
    \caption{Quantitative comparison of the restoration quality of different methods.
            The methods marked in yellow are regression methods, and the methods marked in orange are generative methods.
            The best scores in each category are marked in bold.
            In the case of (a) denoising and (b) deblurring, \MethodName{} records significantly higher PSNR and SSIM scores than the generative methods but slightly lower scores than the regression methods. In LPIPS, \MethodName{} achieves the second best scores. In FID, \MethodName{} achieves the best scores.
            In the case of (c) super-resolution, \MethodName{} reports higher PSNR and SSIM scores with comparable LPIPS and FID scores compared to all the generative models except for GCFSR~\cite{gcfsr}, which is specifically designed for super-resolution.
            Compared to GCFSR, \MethodName{} shows comparable results with a higher PSNR score.
            }
    \label{tab:quantitative_denoising_deblurring}
\vspace{-0.6cm}
\end{table*}

\subsection{Flexible Selection of Regression Methods}
\vspace{-0.1cm}

\MethodName{} supports flexible selection of diverse regression networks.
To validate its flexibility, we employ four different regression networks in the restoration module: UNet~\cite{UNet}, HINet~\cite{hinet}, NAFNet~\cite{NAFNet}, and RRDBNet~\cite{esrgan}.
\Fig{\ref{fig:flexible}} shows the deblurring and super-resolution results of each network and UGPNet equipped with each regression network.
UGPNet successfully brings restoration power from the task-specialized regression networks and combines it with a generative prior. 
Such flexible choice of regression networks allows us to easily equip existing and future regression models with our universal generative prior.

\vspace{-0.1cm}
\begin{figure*}[ht!]
\centering
    \includegraphics[width=2\columnwidth]{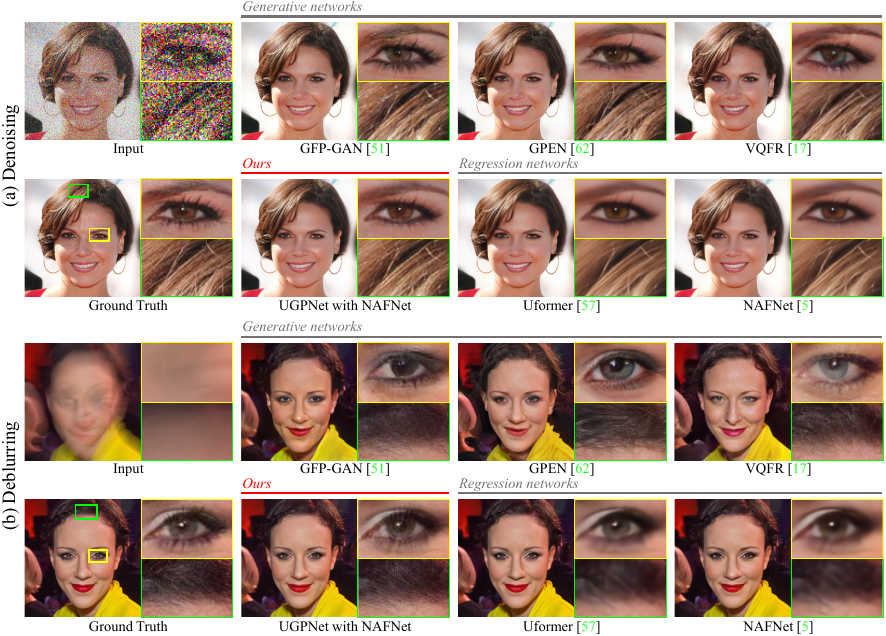}
    \vspace{-0.1cm}
    \caption{
    Qualitative comparison of (a) denoising and (b) deblurring methods: regression methods (Uformer~\cite{uformer_multiple} and NAFNet~\cite{NAFNet}), generative methods (GFP-GAN~\cite{gfpgan}, GPEN~\cite{gpen}, VQFR~\cite{VQFR}), and \MethodName{} with NAFNet~\cite{NAFNet}.
    \MethodName{} recovers authentic image structure and colors compared to generative methods while synthesizing sharp high-frequency details compared to regression methods.
    }
    \vspace{-0.4cm}
\label{fig:denoisedeblur}
\end{figure*}

\begin{figure*}[hbt!]
    \centerline{\includegraphics[width=1.8\columnwidth]{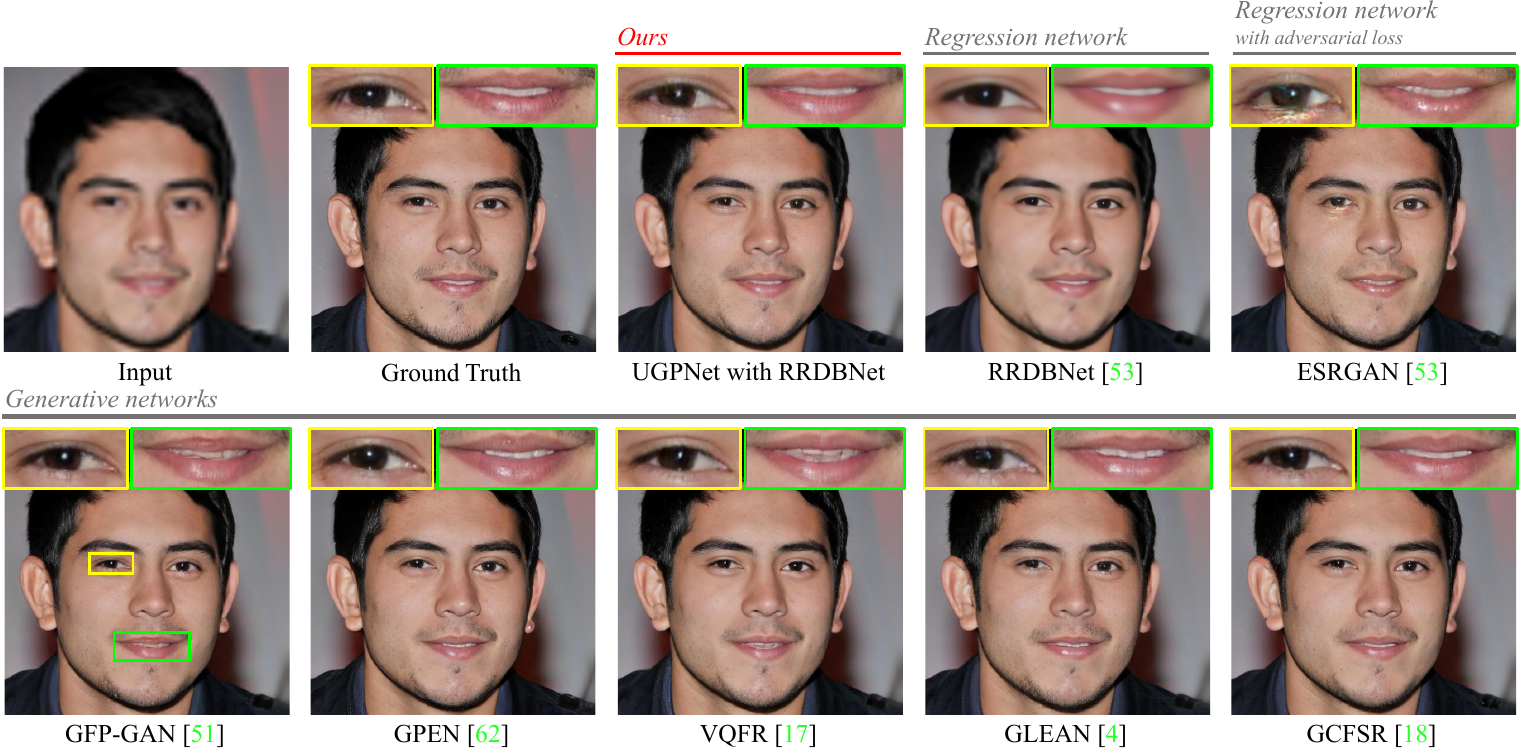}}
    \vspace{-0.1cm}
    \caption{
    Qualitative comparison of super-resolution methods: regression methods (RRDBNet~\cite{esrgan} and ESRGAN~\cite{esrgan}), generative methods (GFP-GAN~\cite{gfpgan}, GPEN~\cite{gpen}, VQFR~\cite{VQFR}, GLEAN~\cite{glean}, GCFSR~\cite{gcfsr}), and \MethodName{} with RRDBNet~\cite{esrgan}.
    \MethodName{} succeeds in generating sharp, realistic high-frequency details compared to regression methods, and shows comparable performance to generative methods.
    }
    \vspace{-0.3cm}
\label{fig:sr}
\end{figure*}

\subsection{Comparison with Restoration Methods}
\vspace{-0.1cm}
We compare \MethodName{} with recent restoration methods whose source codes are publicly released.
Among regression-based methods, we compare \MethodName{} with Uformer~\cite{uformer_multiple} and NAFNet~\cite{NAFNet} for denoising and deblurring, and RRDBNet~\cite{esrgan} for super-resolution.
Among generative-prior-based methods, we compare \MethodName{} with VQFR~\cite{VQFR}, GFP-GAN~\cite{gfpgan} and GPEN~\cite{gpen} for denoising and deblurring. 
We compare two more networks, GLEAN~\cite{glean} and GCFSR~\cite{gcfsr} for super-resolution, both of which are specifically designed for super-resolution.
We train all the methods from scratch on our dataset using the authors' code except GPEN.
For GPEN, we finetuned an officially released model as we found it performs better.

\cref{fig:denoisedeblur} shows qualitative comparisons on denoising and deblurring.
The regression-based methods Uformer~\cite{uformer_multiple} and NAFNet~\cite{NAFNet}) fail to synthesize sharp details compared to the generative-prior-based methods.
The generative methods (GFP-GAN~\cite{gfpgan}, GPEN~\cite{gpen}, and VQFR~\cite{VQFR}) are unable to restore faithful image structures.
\MethodName{} succeeds in the faithful restoration and high-frequency synthesis of image structures and details.
We further report quantitative evaluation in \Tbl{\ref{tab:quantitative_denoising_deblurring}} (a) and (b).
In PSNR and SSIM, \MethodName{} records higher scores than the generative methods, indicating structural consistency.
\MethodName{} has a slightly lower PSNR scores than the best regression methods, due to the synthesized high-frequency details. In LPIPS, \MethodName{} achieves the second best scores. In FID, \MethodName{} achieves the best scores, demonstrating that it generates perceptually-plausible high-frequency details. 

\Fig{\ref{fig:sr}} and \Tbl{\ref{tab:quantitative_denoising_deblurring}} (c) show the qualitative and quantitative comparisons on super-resolution.
As the super-resolution requires more aggressive high-frequency generation, a regression method (RRDBNet~\cite{esrgan}) produces perceptually low-quality blurry images with high FID scores.
Adopting an adversarial loss (ESRGAN~\cite{esrgan}) produces sharp images and lowers the FID score, but fails to synthesize realistic textures as the generative prior-based methods.
Both generative methods and ours succeed in synthesizing realistic high-frequency details showing comparable results thanks to powerful generative priors.
Quantitatively, \MethodName{} achieves higher PSNR and SSIM scores with comparable LPIPS and FID scores compared to all the generative methods except GCFSR~\cite{gcfsr}, which is specifically designed for super-resolution.
Compared to GCFSR, \MethodName{} achieves similar performance, recording a slightly better PSNR score and slightly worse LPIPS and FID scores.
\vspace{-0.5cm}
\vspace{-0.3cm}
\paragraph{Model Complexity}
We compare the number of parameters and the inference speed on deblurring in \cref{tab:complexity}.
Although \MethodName{} combines a regression network (NAFNet~\cite{NAFNet}) and a generative network, it has a comparable number of parameters with the recent models.
This is because the other methods require a lot of parameters in their encoders or decoders, whereas our lightweight model design achieves better performance with fewer parameters.
In terms of inference speed, \MethodName{} is slower than the lightweight models, but it is faster than computationally heavy models such as the transformer-based model~\cite{uformer_multiple} and the dictionary-based model~\cite{VQFR}.
\begin{table}[t!]
\centering
\vspace{-0.1cm}
\setlength{\tabcolsep}{0.3em}
\scalebox{0.7}{
\begin{tabular}{c|c|c|c|c|c}
\hline
\Xhline{3\arrayrulewidth}
                                & NAFNet~\cite{NAFNet} & GFP-GAN~\cite{gfpgan} & VQFR~\cite{VQFR} & Uformer~\cite{uformer_multiple} & UGPNet \\ \hline
Param (M)  &   17.1  &   76.2  & 76.6  & 50.9 &  69.6\\ \hline
Time (ms)  &   45.2   & 22.7    & 181.2 &  170.6 & 90.5 \\ \hline
\Xhline{3\arrayrulewidth}
\end{tabular}
}
\vspace{-0.1cm}
\caption{Comparison of the parameter numbers and the inference times with an NVIDIA GeForce RTX 3090 GPU. \MethodName{} has a comparable number of parameters with the others and it is faster than transformer-based \cite{uformer_multiple} and dictionary-based models \cite{VQFR}. 
}
\label{tab:complexity}
\vspace{-0.2cm}
\end{table}

\begin{figure}[t!]
    \vspace{-0.2cm}
    \centerline{\includegraphics[width=1\columnwidth]{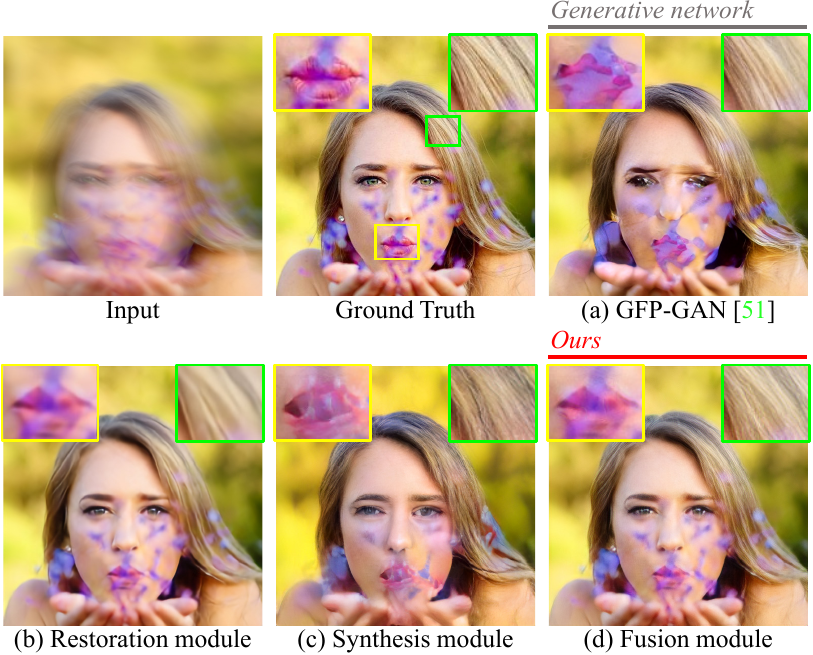}}
    \vspace{-0.15cm}
    \caption{
    \MethodName{} supports a wider range of images compared to the generative methods.
    (a) A generative method fails to synthesize images outside the training distribution and produces artifacts on the face.
    While (c) our synthesis module also introduces artifacts, (d) our fusion module successfully recovers a clean image without artifacts as it selectively uses the synthesized high-frequency details on top of (b) the result of the restoration module.
    }
\vspace{-0.6cm}
\label{fig:extreme_module_result}
\end{figure}

\vspace{-0.4cm}
\paragraph{Restoration of Out-of-Distribution Images}
\MethodName{} is robust against catastrophic failures that generative prior-based methods suffer from when restoring images outside the training distributions, as shown in \cref{fig:teaser} (g) and \cref{fig:extreme_module_result}.
These robustness of \MethodName{} stems from the fusion module that adaptively combines features from the restoration and synthesis modules to achieve faithful and natural-looking restoration.
For a deeper analysis, \cref{fig:extreme_module_result} compares outputs of different modules on a degraded input image outside the training distribution.
In the figure, the restoration module robustly restores the image structures thanks to the well-generalized regression network. 
The synthesis module synthesizes high-frequency details from the output of the restoration module but it also introduces artifacts in the face as the image is outside the training distribution of the generative prior.
Despite such synthesis artifacts, the fusion module selectively uses the synthesized high-frequency details on top of the result of the restoration module and produces the final result that is faithfully reconstructed without artifacts and has natural-looking high-frequency details.

\begin{table}[t!]
\centering
\vspace{-0.1cm}
\setlength{\tabcolsep}{0.3em}
\scalebox{0.7}{
\begin{tabular}{|l|ccc|ccc|}
\hline
\Xhline{4\arrayrulewidth}\multirow{2}{*}{\textit{Configurations}} & \multicolumn{3}{c|}{w/ NAFNet~\cite{NAFNet}}                                                                                   & \multicolumn{3}{c|}{w/ HINet~\cite{hinet}}                                                                                    \\ \cline{2-7} 
                                                         & \multicolumn{1}{c|}{\textbf{PSNR~$\uparrow$}}        & \multicolumn{1}{c|}{\textbf{SSIM~$\uparrow$}}       & \textbf{FID~$\downarrow$}       & \multicolumn{1}{c|}{\textbf{PSNR~$\uparrow$}}        & \multicolumn{1}{c|}{\textbf{SSIM~$\uparrow$}}       & \textbf{FID~$\downarrow$}      \\ \hline
(a) w/o $R_{se}$ and $R_{mg}$                            & \multicolumn{1}{c|}{27.43}                  & \multicolumn{1}{c|}{0.75}                  & 8.33                  & \multicolumn{1}{c|}{27.30}                  & \multicolumn{1}{c|}{0.72}                  & 9.70                  \\ \hline
(b) w/ fusion module                                     & \multicolumn{1}{c|}{\multirow{2}{*}{28.49}} & \multicolumn{1}{c|}{\multirow{2}{*}{\textbf{0.76}}} & \multirow{2}{*}{8.54} & \multicolumn{1}{c|}{\multirow{2}{*}{28.16}} & \multicolumn{1}{c|}{\multirow{2}{*}{0.75}} & \multirow{2}{*}{9.68} \\
combining $x_{reg}$ and $x_{syn}$                        & \multicolumn{1}{c|}{}                       & \multicolumn{1}{c|}{}                      &                       & \multicolumn{1}{c|}{}                       & \multicolumn{1}{c|}{}                      &                       \\ \hline
(c) \MethodName{}                                               & \multicolumn{1}{c|}{\textbf{28.64}}                  & \multicolumn{1}{c|}{\textbf{0.76}}                  & \textbf{8.02}                  & \multicolumn{1}{c|}{\textbf{28.49}}                  & \multicolumn{1}{c|}{\textbf{0.76}}                  & \textbf{8.34}                  \\ \hline\Xhline{4\arrayrulewidth}
\end{tabular}
}
\vspace{-0.1cm}
\caption{
We validate network components of \MethodName{} on deblurring. 
(a) Without the structure encoder and the merging network, the restoration module has difficulty in extracting features with structural information, leading to performance drop. 
(b) \MethodName{} combining images rather than (c) features also leads to slight performance drop.}
\label{tab:ablation}
\vspace{-0.6cm}
\end{table}
\subsection{Ablation Study}

\paragraph{Network Components}
\cref{tab:ablation} quantitatively validates the network components of \MethodName{} on deblurring.
In the restoration module, introducing the structure encoder and the merging network for the residual regression network leads to performance improvement (\Tbl{\ref{tab:ablation}} (a) and (c)).
This improvement demonstrates that conveying additional information from the input helps the restoration feature $f_{reg}$ contain authentic image structures.
In the fusion module, combining the restoration feature $f_{reg}$ and the synthesis feature $f_{syn}$ rather than the image alternatives, $x_{reg}$ and $x_{syn}$, leads to slight performance improvement (\Tbl{\ref{tab:ablation}} (b) and (c)).
\vspace{-0.2cm}
\section{Conclusion}
\label{sec:conclusion}
\vspace{-0.1cm}
This paper proposed \MethodName, a universal generative prior framework for image restoration.
\MethodName{} supports diverse regression networks developed for each task and brings the generative power of recent generative prior on top of it.
Through extensive experiments on deblurring, denoising, and super-resolution, we demonstrated that \MethodName{} succeeds in high-quality image restoration, enabling faithful restoration with realistic high-frequency details.
\vspace{-0.4cm}
\paragraph{Limitation} 

Although \MethodName{} is robust against failure of the generative prior, it may fail if the regression method fails, as shown in \MethodName{} with UNet in \Fig{\ref{fig:flexible}}. 
Also, \MethodName{} is less favorable to PSNR and SSIM scores compared to the regression method, and produces less sharp images than its backbone generative model (StyleGAN2~\cite{stylegan2}), as it aims at higher-fidelity results. 
It would be an interesting future direction to address these challenges, e.g., we may measure the uncertainty of the regression result and use it to adaptively synthesize necessary details for higher fidelity.

\vspace{-0.4cm}

\small
\paragraph{Acknowledgement}
{This work was supported by the NRF grants (No.2023R1A2C200494611, No.2018R1A5A1060031) and IITP grant (No.2019-0-01906, Artificial Intelligence Graduate School Program(POSTECH)) funded by the Korea government (MSIT) and Samsung Electronics Co., Ltd. Seung-Hwan Baek is partly supported by Korea NRF (RS-2023-00211658, 2022R1A6A1A03052954), Korea MOTIE (NTIS1415187366-20025752) and Samsung Research Funding Center (SRFCIT1801-52).}

{\small
\bibliographystyle{ieee_fullname}
\bibliography{egbib}

\begin{thebibliography}{10}\itemsep=-1pt

\bibitem{image2stylegan}
Rameen Abdal, Yipeng Qin, and Peter Wonka.
\newblock Image2stylegan: How to embed images into the stylegan latent space?
\newblock In {\em Proceedings of the IEEE/CVF International Conference on Computer Vision}, pages 4432--4441, 2019.

\bibitem{biggan}
Andrew Brock, Jeff Donahue, and Karen Simonyan.
\newblock Large scale gan training for high fidelity natural image synthesis.
\newblock {\em arXiv preprint arXiv:1809.11096}, 2018.

\bibitem{non-localdenoise}
Antoni Buades, Bartomeu Coll, and J-M Morel.
\newblock A non-local algorithm for image denoising.
\newblock In {\em 2005 IEEE computer society conference on computer vision and pattern recognition (CVPR'05)}, volume~2, pages 60--65. Ieee, 2005.

\bibitem{glean}
Kelvin~CK Chan, Xintao Wang, Xiangyu Xu, Jinwei Gu, and Chen~Change Loy.
\newblock Glean: Generative latent bank for large-factor image super-resolution.
\newblock In {\em Proceedings of the IEEE/CVF conference on computer vision and pattern recognition}, pages 14245--14254, 2021.

\bibitem{NAFNet}
Liangyu Chen, Xiaojie Chu, Xiangyu Zhang, and Jian Sun.
\newblock Simple baselines for image restoration.
\newblock In {\em Computer Vision--ECCV 2022: 17th European Conference, Tel Aviv, Israel, October 23--27, 2022, Proceedings, Part VII}, pages 17--33. Springer, 2022.

\bibitem{hinet}
Liangyu Chen, Xin Lu, Jie Zhang, Xiaojie Chu, and Chengpeng Chen.
\newblock Hinet: Half instance normalization network for image restoration.
\newblock In {\em Proceedings of the IEEE/CVF Conference on Computer Vision and Pattern Recognition}, pages 182--192, 2021.

\bibitem{activating_super}
Xiangyu Chen, Xintao Wang, Jiantao Zhou, and Chao Dong.
\newblock Activating more pixels in image super-resolution transformer.
\newblock {\em arXiv preprint arXiv:2205.04437}, 2022.

\bibitem{nbnet_denoise}
Shen Cheng, Yuzhi Wang, Haibin Huang, Donghao Liu, Haoqiang Fan, and Shuaicheng Liu.
\newblock Nbnet: Noise basis learning for image denoising with subspace projection.
\newblock In {\em Proceedings of the IEEE/CVF Conference on Computer Vision and Pattern Recognition}, pages 4896--4906, 2021.

\bibitem{fastmotiondeblurring}
Sunghyun Cho and Seungyong Lee.
\newblock Fast motion deblurring.
\newblock In {\em ACM SIGGRAPH Asia 2009 papers}, pages 1--8. 2009.

\bibitem{mimounet}
Sung-Jin Cho, Seo-Won Ji, Jun-Pyo Hong, Seung-Won Jung, and Sung-Jea Ko.
\newblock Rethinking coarse-to-fine approach in single image deblurring.
\newblock In {\em Proceedings of the IEEE/CVF international conference on computer vision}, pages 4641--4650, 2021.

\bibitem{collaborativefiltering}
Kostadin Dabov, Alessandro Foi, Vladimir Katkovnik, and Karen Egiazarian.
\newblock Image denoising by sparse 3-d transform-domain collaborative filtering.
\newblock {\em IEEE Transactions on image processing}, 16(8):2080--2095, 2007.

\bibitem{wavelet2}
Xin Deng, Ren Yang, Mai Xu, and Pier~Luigi Dragotti.
\newblock Wavelet domain style transfer for an effective perception-distortion tradeoff in single image super-resolution.
\newblock In {\em Proceedings of the IEEE/CVF international conference on computer vision}, pages 3076--3085, 2019.

\bibitem{SRCNN}
Chao Dong, Chen~Change Loy, Kaiming He, and Xiaoou Tang.
\newblock Learning a deep convolutional network for image super-resolution.
\newblock In {\em European conference on computer vision}, pages 184--199. Springer, 2014.

\bibitem{vqgan}
Patrick Esser, Robin Rombach, and Bjorn Ommer.
\newblock Taming transformers for high-resolution image synthesis.
\newblock In {\em Proceedings of the IEEE/CVF conference on computer vision and pattern recognition}, pages 12873--12883, 2021.

\bibitem{removingshake}
Rob Fergus, Barun Singh, Aaron Hertzmann, Sam~T Roweis, and William~T Freeman.
\newblock Removing camera shake from a single photograph.
\newblock In {\em Acm Siggraph 2006 Papers}, pages 787--794. 2006.

\bibitem{mganprior}
Jinjin Gu, Yujun Shen, and Bolei Zhou.
\newblock Image processing using multi-code gan prior.
\newblock In {\em Proceedings of the IEEE/CVF conference on computer vision and pattern recognition}, pages 3012--3021, 2020.

\bibitem{VQFR}
Yuchao Gu, Xintao Wang, Liangbin Xie, Chao Dong, Gen Li, Ying Shan, and Ming-Ming Cheng.
\newblock Vqfr: Blind face restoration with vector-quantized dictionary and parallel decoder.
\newblock In {\em Computer Vision--ECCV 2022: 17th European Conference, Tel Aviv, Israel, October 23--27, 2022, Proceedings, Part XVIII}, pages 126--143. Springer, 2022.

\bibitem{gcfsr}
Jingwen He, Wu Shi, Kai Chen, Lean Fu, and Chao Dong.
\newblock Gcfsr: a generative and controllable face super resolution method without facial and gan priors.
\newblock In {\em Proceedings of the IEEE/CVF Conference on Computer Vision and Pattern Recognition}, pages 1889--1898, 2022.

\bibitem{fid}
Martin Heusel, Hubert Ramsauer, Thomas Unterthiner, Bernhard Nessler, and Sepp Hochreiter.
\newblock Gans trained by a two time-scale update rule converge to a local nash equilibrium.
\newblock In I. Guyon, U.~Von Luxburg, S. Bengio, H. Wallach, R. Fergus, S. Vishwanathan, and R. Garnett, editors, {\em Advances in Neural Information Processing Systems}, volume~30, 2017.

\bibitem{DDPM}
Jonathan Ho, Ajay Jain, and Pieter Abbeel.
\newblock Denoising diffusion probabilistic models.
\newblock {\em Advances in Neural Information Processing Systems}, 33:6840--6851, 2020.

\bibitem{pyramid_deblur}
Xiaobin Hu, Wenqi Ren, Kaicheng Yu, Kaihao Zhang, Xiaochun Cao, Wei Liu, and Bjoern Menze.
\newblock Pyramid architecture search for real-time image deblurring.
\newblock In {\em Proceedings of the IEEE/CVF International Conference on Computer Vision}, pages 4298--4307, 2021.

\bibitem{BDInvert}
Kyoungkook Kang, Seongtae Kim, and Sunghyun Cho.
\newblock Gan inversion for out-of-range images with geometric transformations.
\newblock In {\em Proceedings of the IEEE/CVF International Conference on Computer Vision}, pages 13941--13949, 2021.

\bibitem{pggan}
Tero Karras, Timo Aila, Samuli Laine, and Jaakko Lehtinen.
\newblock Progressive growing of gans for improved quality, stability, and variation.
\newblock {\em arXiv preprint arXiv:1710.10196}, 2017.

\bibitem{stylegan1}
Tero Karras, Samuli Laine, and Timo Aila.
\newblock A style-based generator architecture for generative adversarial networks.
\newblock In {\em Proceedings of the IEEE/CVF conference on computer vision and pattern recognition}, pages 4401--4410, 2019.

\bibitem{stylegan2}
Tero Karras, Samuli Laine, Miika Aittala, Janne Hellsten, Jaakko Lehtinen, and Timo Aila.
\newblock Analyzing and improving the image quality of stylegan.
\newblock In {\em Proceedings of the IEEE/CVF conference on computer vision and pattern recognition}, pages 8110--8119, 2020.

\bibitem{DDRM}
Bahjat Kawar, Michael Elad, Stefano Ermon, and Jiaming Song.
\newblock Denoising diffusion restoration models.
\newblock {\em arXiv preprint arXiv:2201.11793}, 2022.

\bibitem{bigcolor}
Geonung Kim, Kyoungkook Kang, Seongtae Kim, Hwayoon Lee, Sehoon Kim, Jonghyun Kim, Seung-Hwan Baek, and Sunghyun Cho.
\newblock Bigcolor: Colorization using a generative color prior for natural images.
\newblock In {\em Computer Vision--ECCV 2022: 17th European Conference, Tel Aviv, Israel, October 23--27, 2022, Proceedings, Part VII}, pages 350--366. Springer, 2022.

\bibitem{vdsr_residual}
Jiwon Kim, Jung~Kwon Lee, and Kyoung~Mu Lee.
\newblock Accurate image super-resolution using very deep convolutional networks.
\newblock In {\em Proceedings of the IEEE conference on computer vision and pattern recognition}, pages 1646--1654, 2016.

\bibitem{mssnet_deblur}
Kiyeon Kim, Seungyong Lee, and Sunghyun Cho.
\newblock Mssnet: Multi-scale-stage network for single image deblurring.
\newblock {\em arXiv preprint arXiv:2202.09652}, 2022.

\bibitem{transfer_denoise}
Yoonsik Kim, Jae~Woong Soh, Gu~Yong Park, and Nam~Ik Cho.
\newblock Transfer learning from synthetic to real-noise denoising with adaptive instance normalization.
\newblock In {\em Proceedings of the IEEE/CVF Conference on Computer Vision and Pattern Recognition}, pages 3482--3492, 2020.

\bibitem{deblurgan}
Orest Kupyn, Volodymyr Budzan, Mykola Mykhailych, Dmytro Mishkin, and Ji{\v{r}}{\'\i} Matas.
\newblock Deblurgan: Blind motion deblurring using conditional adversarial networks.
\newblock In {\em Proceedings of the IEEE conference on computer vision and pattern recognition}, pages 8183--8192, 2018.

\bibitem{deblurgan2}
Orest Kupyn, Tetiana Martyniuk, Junru Wu, and Zhangyang Wang.
\newblock Deblurgan-v2: Deblurring (orders-of-magnitude) faster and better.
\newblock In {\em Proceedings of the IEEE/CVF International Conference on Computer Vision}, pages 8878--8887, 2019.

\bibitem{srgan}
Christian Ledig, Lucas Theis, Ferenc Husz{\'a}r, Jose Caballero, Andrew Cunningham, Alejandro Acosta, Andrew Aitken, Alykhan Tejani, Johannes Totz, Zehan Wang, et~al.
\newblock Photo-realistic single image super-resolution using a generative adversarial network.
\newblock In {\em Proceedings of the IEEE conference on computer vision and pattern recognition}, pages 4681--4690, 2017.

\bibitem{Blindspot_denoise}
Wooseok Lee, Sanghyun Son, and Kyoung~Mu Lee.
\newblock Ap-bsn: Self-supervised denoising for real-world images via asymmetric pd and blind-spot network.
\newblock In {\em Proceedings of the IEEE/CVF Conference on Computer Vision and Pattern Recognition}, pages 17725--17734, 2022.

\bibitem{invertible_denoise}
Yang Liu, Zhenyue Qin, Saeed Anwar, Pan Ji, Dongwoo Kim, Sabrina Caldwell, and Tom Gedeon.
\newblock Invertible denoising network: A light solution for real noise removal.
\newblock In {\em Proceedings of the IEEE/CVF conference on computer vision and pattern recognition}, pages 13365--13374, 2021.

\bibitem{progressive_super}
Yuqing Liu, Xinfeng Zhang, Shanshe Wang, Siwei Ma, and Wen Gao.
\newblock Progressive multi-scale residual network for single image super-resolution.
\newblock {\em arXiv preprint arXiv:2007.09552}, 2020.

\bibitem{Fourier_deblur}
Xintian Mao, Yiming Liu, Wei Shen, Qingli Li, and Yan Wang.
\newblock Deep residual fourier transformation for single image deblurring.
\newblock {\em arXiv preprint arXiv:2111.11745}, 2021.

\bibitem{contextual}
Roey Mechrez, Itamar Talmi, and Lihi Zelnik-Manor.
\newblock The contextual loss for image transformation with non-aligned data.
\newblock In {\em Proceedings of the European conference on computer vision (ECCV)}, pages 768--783, 2018.

\bibitem{mining_super}
Yiqun Mei, Yuchen Fan, Yuqian Zhou, Lichao Huang, Thomas~S Huang, and Honghui Shi.
\newblock Image super-resolution with cross-scale non-local attention and exhaustive self-exemplars mining.
\newblock In {\em Proceedings of the IEEE/CVF conference on computer vision and pattern recognition}, pages 5690--5699, 2020.

\bibitem{pulse}
Sachit Menon, Alexandru Damian, Shijia Hu, Nikhil Ravi, and Cynthia Rudin.
\newblock Pulse: Self-supervised photo upsampling via latent space exploration of generative models.
\newblock In {\em Proceedings of the ieee/cvf conference on computer vision and pattern recognition}, pages 2437--2445, 2020.

\bibitem{Gopro}
Seungjun Nah, Tae Hyun~Kim, and Kyoung Mu~Lee.
\newblock Deep multi-scale convolutional neural network for dynamic scene deblurring.
\newblock In {\em Proceedings of the IEEE conference on computer vision and pattern recognition}, pages 3883--3891, 2017.

\bibitem{holistic_super}
Ben Niu, Weilei Wen, Wenqi Ren, Xiangde Zhang, Lianping Yang, Shuzhen Wang, Kaihao Zhang, Xiaochun Cao, and Haifeng Shen.
\newblock Single image super-resolution via a holistic attention network.
\newblock In {\em European conference on computer vision}, pages 191--207. Springer, 2020.

\bibitem{pSp}
Elad Richardson, Yuval Alaluf, Or Patashnik, Yotam Nitzan, Yaniv Azar, Stav Shapiro, and Daniel Cohen-Or.
\newblock Encoding in style: a stylegan encoder for image-to-image translation.
\newblock In {\em Proceedings of the IEEE/CVF conference on computer vision and pattern recognition}, pages 2287--2296, 2021.

\bibitem{RealBlur}
Jaesung Rim, Haeyun Lee, Jucheol Won, and Sunghyun Cho.
\newblock Real-world blur dataset for learning and benchmarking deblurring algorithms.
\newblock In {\em European Conference on Computer Vision}, pages 184--201. Springer, 2020.

\bibitem{UNet}
Olaf Ronneberger, Philipp Fischer, and Thomas Brox.
\newblock U-net: Convolutional networks for biomedical image segmentation.
\newblock In {\em International Conference on Medical image computing and computer-assisted intervention}, pages 234--241. Springer, 2015.

\bibitem{diffusion_sr}
Chitwan Saharia, Jonathan Ho, William Chan, Tim Salimans, David~J Fleet, and Mohammad Norouzi.
\newblock Image super-resolution via iterative refinement.
\newblock {\em IEEE Transactions on Pattern Analysis and Machine Intelligence}, 2022.

\bibitem{attentive_deblur}
Maitreya Suin, Kuldeep Purohit, and AN Rajagopalan.
\newblock Spatially-attentive patch-hierarchical network for adaptive motion deblurring.
\newblock In {\em Proceedings of the IEEE/CVF Conference on Computer Vision and Pattern Recognition}, pages 3606--3615, 2020.

\bibitem{SRN}
Xin Tao, Hongyun Gao, Xiaoyong Shen, Jue Wang, and Jiaya Jia.
\newblock Scale-recurrent network for deep image deblurring.
\newblock In {\em Proceedings of the IEEE conference on computer vision and pattern recognition}, pages 8174--8182, 2018.

\bibitem{e4e}
Omer Tov, Yuval Alaluf, Yotam Nitzan, Or Patashnik, and Daniel Cohen-Or.
\newblock Designing an encoder for stylegan image manipulation.
\newblock {\em ACM Transactions on Graphics (TOG)}, 40(4):1--14, 2021.

\bibitem{maxim_multiple}
Zhengzhong Tu, Hossein Talebi, Han Zhang, Feng Yang, Peyman Milanfar, Alan Bovik, and Yinxiao Li.
\newblock Maxim: Multi-axis mlp for image processing.
\newblock In {\em Proceedings of the IEEE/CVF Conference on Computer Vision and Pattern Recognition}, pages 5769--5780, 2022.

\bibitem{gfpgan}
Xintao Wang, Yu Li, Honglun Zhang, and Ying Shan.
\newblock Towards real-world blind face restoration with generative facial prior.
\newblock In {\em Proceedings of the IEEE/CVF Conference on Computer Vision and Pattern Recognition}, pages 9168--9178, 2021.

\bibitem{realesrgan}
Xintao Wang, Liangbin Xie, Chao Dong, and Ying Shan.
\newblock Real-esrgan: Training real-world blind super-resolution with pure synthetic data.
\newblock In {\em Proceedings of the IEEE/CVF International Conference on Computer Vision}, pages 1905--1914, 2021.

\bibitem{esrgan}
Xintao Wang, Ke Yu, Shixiang Wu, Jinjin Gu, Yihao Liu, Chao Dong, Yu Qiao, and Chen Change~Loy.
\newblock Esrgan: Enhanced super-resolution generative adversarial networks.
\newblock In {\em Proceedings of the European conference on computer vision (ECCV) workshops}, pages 0--0, 2018.

\bibitem{practical_denoise}
Yuzhi Wang, Haibin Huang, Qin Xu, Jiaming Liu, Yiqun Liu, and Jue Wang.
\newblock Practical deep raw image denoising on mobile devices.
\newblock In {\em European Conference on Computer Vision}, pages 1--16. Springer, 2020.

\bibitem{DDNM}
Yinhuai Wang, Jiwen Yu, and Jian Zhang.
\newblock Zero-shot image restoration using denoising diffusion null-space model.
\newblock {\em arXiv preprint arXiv:2212.00490}, 2022.

\bibitem{ssim}
Zhou Wang, A.C. Bovik, H.R. Sheikh, and E.P. Simoncelli.
\newblock Image quality assessment: from error visibility to structural similarity.
\newblock {\em IEEE Transactions on Image Processing}, 13(4):600--612, 2004.

\bibitem{uformer_multiple}
Zhendong Wang, Xiaodong Cun, Jianmin Bao, Wengang Zhou, Jianzhuang Liu, and Houqiang Li.
\newblock Uformer: A general u-shaped transformer for image restoration.
\newblock In {\em Proceedings of the IEEE/CVF Conference on Computer Vision and Pattern Recognition}, pages 17683--17693, 2022.

\bibitem{diffusion_deblur}
Jay Whang, Mauricio Delbracio, Hossein Talebi, Chitwan Saharia, Alexandros~G Dimakis, and Peyman Milanfar.
\newblock Deblurring via stochastic refinement.
\newblock In {\em Proceedings of the IEEE/CVF Conference on Computer Vision and Pattern Recognition}, pages 16293--16303, 2022.

\bibitem{towardvivid}
Yanze Wu, Xintao Wang, Yu Li, Honglun Zhang, Xun Zhao, and Ying Shan.
\newblock Towards vivid and diverse image colorization with generative color prior.
\newblock In {\em Proceedings of the IEEE/CVF International Conference on Computer Vision}, pages 14377--14386, 2021.

\bibitem{wavelet1}
Jun Xiao, Tianshan Liu, Rui Zhao, and Kin-Man Lam.
\newblock Balanced distortion and perception in single-image super-resolution based on optimal transport in wavelet domain.
\newblock {\em Neurocomputing}, 464:408--420, 2021.

\bibitem{motionaware_deblur}
Dan Yang and Mehmet Yamac.
\newblock Motion aware double attention network for dynamic scene deblurring.
\newblock In {\em Proceedings of the IEEE/CVF Conference on Computer Vision and Pattern Recognition}, pages 1113--1123, 2022.

\bibitem{gpen}
Tao Yang, Peiran Ren, Xuansong Xie, and Lei Zhang.
\newblock Gan prior embedded network for blind face restoration in the wild.
\newblock In {\em Proceedings of the IEEE/CVF Conference on Computer Vision and Pattern Recognition}, pages 672--681, 2021.

\bibitem{lsun}
Fisher Yu, Ari Seff, Yinda Zhang, Shuran Song, Thomas Funkhouser, and Jianxiong Xiao.
\newblock Lsun: Construction of a large-scale image dataset using deep learning with humans in the loop.
\newblock {\em arXiv preprint arXiv:1506.03365}, 2015.

\bibitem{VDN}
Zongsheng Yue, Hongwei Yong, Qian Zhao, Deyu Meng, and Lei Zhang.
\newblock Variational denoising network: Toward blind noise modeling and removal.
\newblock {\em Advances in neural information processing systems}, 32, 2019.

\bibitem{restormer_multiple}
Syed~Waqas Zamir, Aditya Arora, Salman Khan, Munawar Hayat, Fahad~Shahbaz Khan, and Ming-Hsuan Yang.
\newblock Restormer: Efficient transformer for high-resolution image restoration.
\newblock In {\em Proceedings of the IEEE/CVF Conference on Computer Vision and Pattern Recognition}, pages 5728--5739, 2022.

\bibitem{MIRNet_multiple}
Syed~Waqas Zamir, Aditya Arora, Salman Khan, Munawar Hayat, Fahad~Shahbaz Khan, Ming-Hsuan Yang, and Ling Shao.
\newblock Learning enriched features for real image restoration and enhancement.
\newblock In {\em European Conference on Computer Vision}, pages 492--511. Springer, 2020.

\bibitem{mprnet_multiple}
Syed~Waqas Zamir, Aditya Arora, Salman Khan, Munawar Hayat, Fahad~Shahbaz Khan, Ming-Hsuan Yang, and Ling Shao.
\newblock Multi-stage progressive image restoration.
\newblock In {\em Proceedings of the IEEE/CVF conference on computer vision and pattern recognition}, pages 14821--14831, 2021.

\bibitem{DnCNN}
Kai Zhang, Wangmeng Zuo, Yunjin Chen, Deyu Meng, and Lei Zhang.
\newblock Beyond a gaussian denoiser: Residual learning of deep cnn for image denoising.
\newblock {\em IEEE transactions on image processing}, 26(7):3142--3155, 2017.

\bibitem{FFDNet}
Kai Zhang, Wangmeng Zuo, and Lei Zhang.
\newblock Ffdnet: Toward a fast and flexible solution for cnn-based image denoising.
\newblock {\em IEEE Transactions on Image Processing}, 27(9):4608--4622, 2018.

\bibitem{lpips}
Richard Zhang, Phillip Isola, Alexei~A Efros, Eli Shechtman, and Oliver Wang.
\newblock The unreasonable effectiveness of deep features as a perceptual metric.
\newblock In {\em Proceedings of the IEEE conference on computer vision and pattern recognition}, pages 586--595, 2018.

\bibitem{RCAN}
Yulun Zhang, Kunpeng Li, Kai Li, Lichen Wang, Bineng Zhong, and Yun Fu.
\newblock Image super-resolution using very deep residual channel attention networks.
\newblock In {\em Proceedings of the European conference on computer vision (ECCV)}, pages 286--301, 2018.

\bibitem{context_super}
Yulun Zhang, Donglai Wei, Can Qin, Huan Wang, Hanspeter Pfister, and Yun Fu.
\newblock Context reasoning attention network for image super-resolution.
\newblock In {\em Proceedings of the IEEE/CVF International Conference on Computer Vision}, pages 4278--4287, 2021.

\bibitem{sgpn}
Feida Zhu, Junwei Zhu, Wenqing Chu, Xinyi Zhang, Xiaozhong Ji, Chengjie Wang, and Ying Tai.
\newblock Blind face restoration via integrating face shape and generative priors.
\newblock In {\em Proceedings of the IEEE/CVF Conference on Computer Vision and Pattern Recognition}, pages 7662--7671, 2022.

\end{thebibliography}
}

\end{document}